\documentclass[11pt,letterpaper]{article}
\usepackage{color}
\usepackage[dvipsnames]{xcolor}
\usepackage{ijcnlp2017}
\usepackage{times}
\usepackage{latexsym}

\usepackage{graphicx}
\usepackage{amsmath}
\usepackage{tabularx}
\newcommand{\twolinecell}[2][l]{\begin{tabular}[#1]{@{}l@{}}#2\end{tabular}}
\usepackage{multirow}
\usepackage{hhline}

\usepackage{xparse}

\NewDocumentCommand{\todo}{ mO{} }{\textcolor{OrangeRed}{\textsuperscript{\textit{#2}}\textsf{\textbf{\small[[#1]]}}}}
\NewDocumentCommand{\done}{ mO{} }{\textcolor{BlueGreen}{\textsuperscript{\textit{#1}}\textsf{\textbf{\small[Done]}}}}

\usepackage{caption}

\ijcnlpfinalcopy

\title{Acquiring Background Knowledge to Improve Moral Value Prediction}

\author{Ying Lin$^1$ , Joe Hoover$^2$, Morteza Dehghani$^{2,3}$, Marlon Mooijman$^2$, Heng Ji$^1$\\
  $^1$ Computer Science Department,\\
  Rensselaer Polytechnic Institute, Troy, NY, USA \\
  \texttt{\{liny9,jih\}@rpi.edu} \\
  $^2$ Department of Psychology, \\
  University of Southern California, Los Angeles, CA, USA \\
  \texttt{\{jehoover,mdehghan,mooijman\}@usc.edu} \\
  $^3$ Department of Computer Science, \\
  University of Southern California, Los Angeles, CA, USA
  }

\date{}

\begin{document}

\maketitle

\begin{abstract}

In this paper, we address the problem of detecting expressions of moral values in tweets using content analysis. This is a particularly challenging problem because moral values are often only implicitly signaled in language, and tweets contain little contextual information due to length constraints. To address these obstacles, we present a novel approach to automatically acquire background knowledge from an external knowledge base to enrich input texts and thus improve moral value prediction. By combining basic text features with background knowledge, our overall context-aware framework achieves performance comparable to a single human annotator. To the best of our knowledge, this is the first attempt to incorporate background knowledge for the prediction of implicit psychological variables in the area of computational social science.

\end{abstract}

\section{Introduction}

\begin{figure}[!hbt]
\centering
\includegraphics[width=7.8cm]{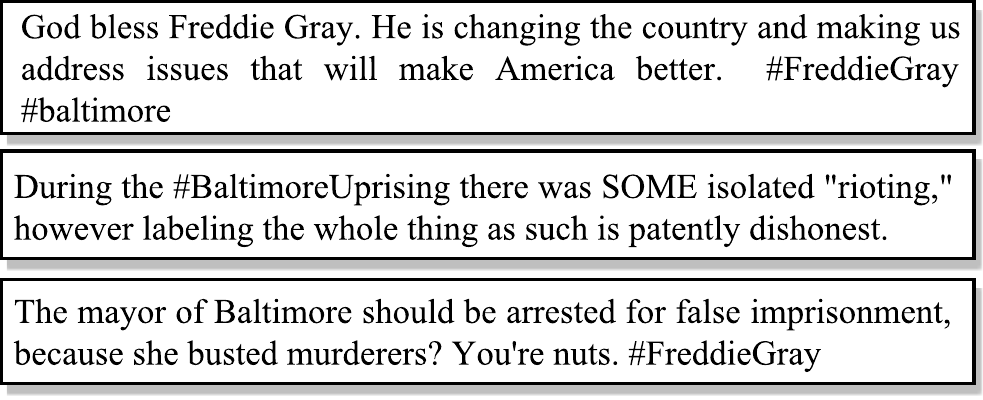}
\caption{Tweets related to 2015 Baltimore Protest.}
\label{figure:baltimore_examples}
\end{figure}

Moral values are principles that define right and wrong for a given individual. They influence decision making, social judgments, motivation, and behavior and are thought of as the glue that holds society together \cite{haidt2012righteous}. However, moral values are not universal, and disagreements about what is moral or sacred can give rise to seemingly intractable conflicts \cite{dehghani2010sacred,ginges2007sacred}.
Accordingly, public demonstrations and protests often involve moral conflicts between different groups. For example, as Figure~\ref{figure:baltimore_examples} shows, during the 2015 Baltimore protests\footnote{https://en.wikipedia.org/wiki/2015\_Baltimore\_protests}, users posted their viewpoints about this event on Twitter, demonstrating divergent and even opposite moral values.

Detecting moral values in user-generated content not only can provide insight into these conflicts but also inform applications that aim to model social phenomena such as voting behavior and public opinions. For example, \cite{koleva2012tracing} shows that moral concerns play an important role in one's attitude and ideological position across a wide range of issues, such abortion and same-sex marriage. Moral values have also been used to investigate various political attitudes in the United States. Liberals and conservatives attend to different moral intuitions~\cite{graham2009liberals}: Liberals focus on the notions of Harm and Fairness, while conservatives attend to ideas of Loyalty to in-group members, Authority, and Purity.

\begin{figure}[!hbt]
\centering
\includegraphics[width=7.5cm]{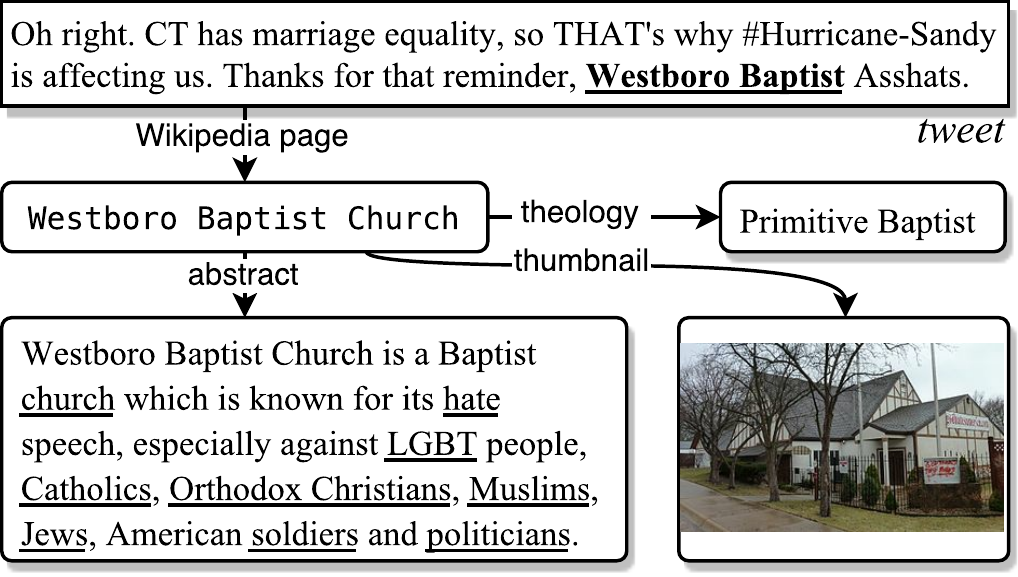}
\caption{Example of Westboro Baptist Church.}
\label{figure:wbc_example}
\end{figure}

In this work, we predict the moral values expressed in social media text via a suite of Natural Language Processing (NLP) techniques.
A given text can contain any one or more moral values, as defined by Moral Foundation Theory (MFT, elaborated in Section~\ref{section:mft})~\cite{gramham2013mft}, or it can be \textit{non-moral}. In previous work, computational linguistic measurements of latent attributes such as moral values, personality, and political orientation have primarily relied on textual features directly derived from target texts; these features have ranged from $n$-grams, word embeddings, emoticons, to word categories~\cite{rao2010classifying,tumasjan2010predicting,golbeck2011predicting,conover2011political,schwartz2013personality,dehghani2014analyzing,dehghani2016purity}. While such approaches can yield powerful representations of text, they fall far short of human representation, which is greatly enhanced by the capacity to actively acquire background knowledge for reasoning and prediction. In the domain of moral value detection, the capacity for external knowledge integration is particularly important.
For example, consider the tweet shown in Figure~\ref{figure:wbc_example}. A reader who has no knowledge of ``Westboro Baptist'' could look it up and learn that it is a church known for anti-LGBT and racist hate speech. This reader might then infer that this tweet conveys moral values concerning Purity/Degradation and Fairness/Cheating. Conversely, an algorithm that lacks access to background knowledge would be unable to exploit this information-rich indicator. Accordingly, we apply Entity Linking (EL) to identify entities in tweets, link them to an external knowledge-base (KB; Wikipedia in this work), and acquire their abstract descriptions and properties. From the background knowledge, we extract words showing a strong correlation with each moral foundation as additional discriminative features to improve the prediction.

Overall, this paper makes the following contributions:

1. We introduce various NLP techniques, such as entity linking and part-of-speech tagging, to tackle the problem of moral value prediction, which provides a new insight into the inference of latent semantic attributes in social media.

2. In the area of computational social science, most previous work involving applications of NLP to psychological measurement has relied exclusively on features derived directly from input text. Due to the brevity and informality of tweets, however, textual features alone may not be sufficient for high-quality prediction. To address this issue, we acquire and incorporate background knowledge into our language models in order to better represent tweets and we use moral value prediction as case study for this approach. To the best of our knowledge, this is the first work to actively acquire background knowledge to enrich contextual information for more precise prediction in computational social science applications.

\section{Moral Foundation Theory}
\label{section:mft}

What a given person holds to be moral or immoral can vary widely as a function of individual differences, and contextual and cultural factors. Moral Foundations Theory~\cite{gramham2013mft}~\footnote{http://moralfoundations.org/} aims to explain this variability as a function of five core moral factors or foundations that appear across cultures, as shown in Table~\ref{table:moral_foundations}. These foundations account for various aspects of morality that serve different but related social functions. Degree of sensitivity towards them vary across different cultures and can change over time.

\begin{table}[!htb]
\small
\centering
\setlength\tabcolsep{2pt}
\begin{tabularx}{0.48\textwidth}{|l|X|}
\hline
\textbf{Foundation} & \textbf{Definition} \\
\hline
\twolinecell[t]{Care\\Harm} & Prescriptive moral values such as caring for others, generosity and compassion and moral values prohibiting actions that harm others. \\
\hline
\twolinecell[t]{Fairness\\Cheating} & Prescriptive moral values such as fairness, justice, and reciprocity and moral values prohibiting cheating.\\
\hline
\twolinecell[t]{Loyalty\\Betrayal} & Prescriptive moral values associated with group affiliation and solidarity and moral values prohibiting betrayal of one's group.\\
\hline
\twolinecell[t]{Authority\\Subversion} & Prescriptive moral values associated with fulfilling social roles and submitting to authority and moral values prohibiting rebellion against authority.\\
\hline
\twolinecell[t]{Purity\\Degradation} & Prescriptive moral values associated with  the sacred and holy and moral values prohibiting violating the sacred.\\
\hline
\end{tabularx}
\caption{Moral foundation definitions.}
\label{table:moral_foundations}
\end{table}

Given the importance of human morality for social functioning \cite{haidt2012righteous}, it is perhaps unsurprising that our moral values leave residue in cultural artifacts such as texts. Indeed, research indicates that variation in moral rhetoric can reliably distinguish between cultural groups \cite{graham2009liberals}, is responsive to environmental disturbances such as terrorism \cite{Sagi2014qf}, and predicts psychologically relevant behavior \cite{dehghani2016purity}. 

While classifying the ground-truth moral content of a text is ultimately subjective and imperfect, general sentiment associated with the foundations above has been shown to be a sufficient proxy for models making secondary predictions \cite{graham2009liberals,Sagi2014qf,garten2016morality,dehghani2016purity}.

In Table~\ref{table:moral_foundation_example}, we list real tweets on the topic of Hurricane Sandy extracted from our data set that reflect each of the five foundations.

\begin{table}[!htb]
\small
\centering
\setlength\tabcolsep{3pt}
\begin{tabularx}{0.48\textwidth}{|l|X|}
\hline
\textbf{Foundation} & \textbf{Example} \\
\hline
\twolinecell[t]{Care\\Harm} & Loss of material things \textbf{hurts} but \textbf{loss} of people and pets is \textbf{devastating} Sending \textbf{prayers} to all who were affected by Sandy \\ \hline
\twolinecell[t]{Fairness\\Cheating} & Complicit lap dog \textbf{biased} corrupt media is saying Obama has done good job w Sandy WHAT \textbf{LIES} Organization \& Distribution get double F s \\ \hline
\twolinecell[t]{Loyalty\\Betrayal} & Love my \textbf{fellow brothers and sisters} in New Jeersey [sic] And \textbf{fellow Americans} standing strong as a nation Sandy please donate to local shelters \\ \hline
\twolinecell[t]{Authority\\Subversion} & I maintain a profound \textbf{respect for govchristie} newjersey sandy AT\_USER humanitarian \\ \hline
\twolinecell[t]{Purity\\Degradation} & \textbf{God bless} these men Truly touched by their dedication AT\_USER gentTN Tomb guards Incredible Sandy \\
\hline
\end{tabularx}
\caption{Tweets reflecting each of the foundations.}
\label{table:moral_foundation_example}
\end{table}

\section{Approach Overview}

In this work, our goal is to predict the moral values expressed in social media text based on the Moral Foundation Theory via a suite of Natural Language Processing techniques. For example, moral values on Care/Harm and Purity/Degradation are expected to be detected from the following tweet -- ``\textit{The Lord our Shepherd will keep \& protect everyone on the East Coast Apply wisdom \& be safe. Listen to the Spirit’s nudge. Love you. \#Sandy}''.
Thus, we define the Moral Value Prediction problem as follows:

\textsc{Definition}: \textit{Given a set of documents $\mathcal{X}=\lbrace x_1, ..., x_n\rbrace$ regarding a certain topic and a set of moral foundations $\mathcal{F}=\lbrace f_1, ..., f_m\rbrace$, for each $x \in \mathcal{X}$, return a binary vector $\textbf{y}=\lbrace y_1, ..., y_m\rbrace$, where $y_j$ indicates whether $x$ reflects concern on $f_j$.}

\subsection{Framework}

\begin{figure}[!hbt]
\centering
\includegraphics[width=7.5cm]{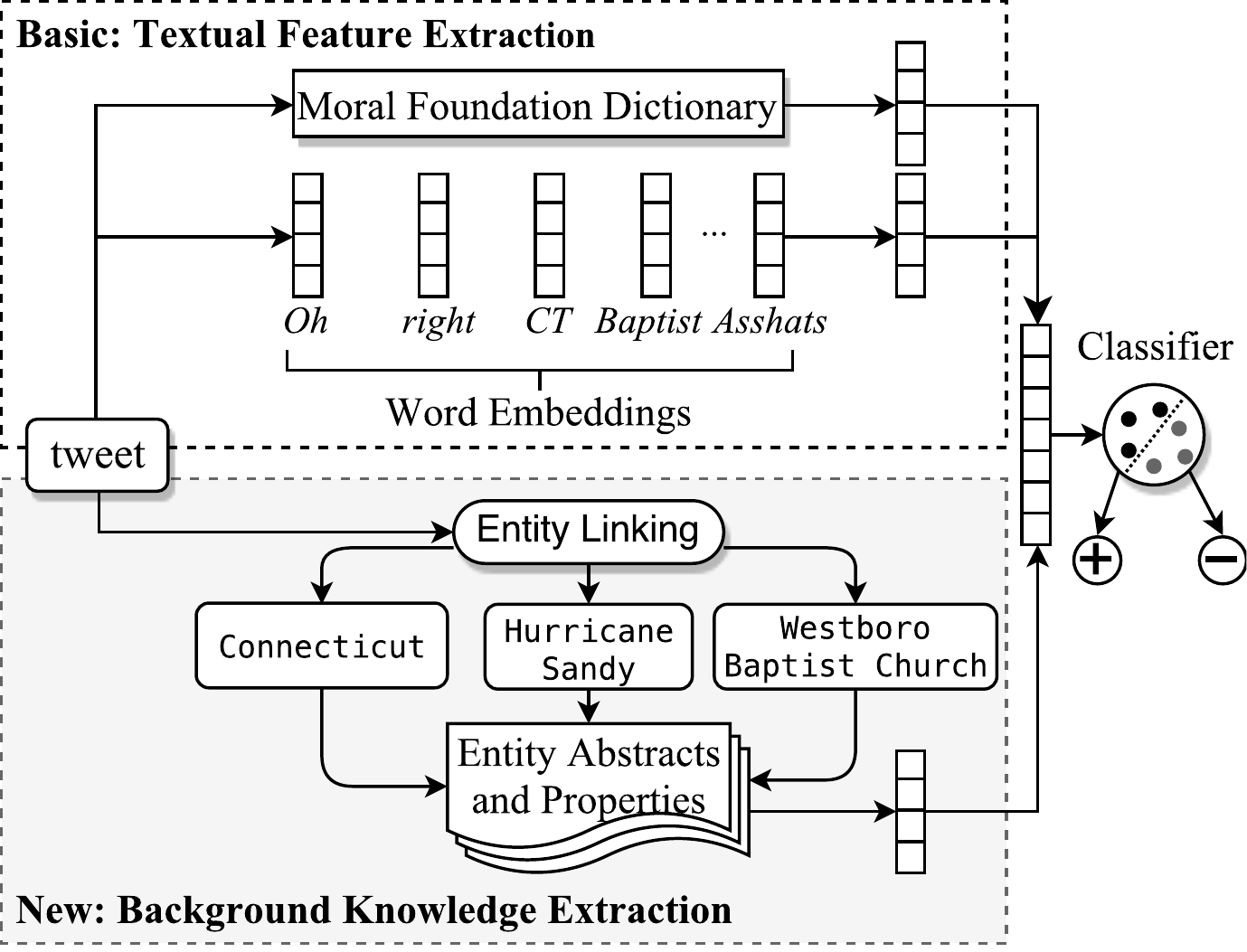}
\caption{Overall framework.}
\label{figure:overall_framework}
\end{figure}

Figure~\ref{figure:overall_framework} depicts the overall framework. In the \textit{Textual Feature Extraction} module, textual features are extracted from the tweet and encoded into separate vectors. At the beginning of the \textit{Background Knowledge Extraction} module, we apply entity linking to each tweet and acquire abstract descriptions and properties of linked entities from the KB. Next, we select information that is relevant to the target moral foundation from the prior knowledge and represent it as a vector. Lastly, all vectors are concatenated as input to a binary classifier.

We train a separate classifier that returns $y_j$ for each foundation $f_j$ and merge classification results from all classifiers as $\textbf{y}$. A tweet will be predicted as ``Non-moral'' if all classifiers return \texttt{False}.

\subsection{Learning Model}

In previous studies on predicting attributes such as gender, personality, power, and political orientation~\cite{gomez2003centrality,burger2011discriminating,schwartz2013personality,park2015automatic,KR2016},
a document is usually modeled as a bag of words and represented by counting the frequency of each feature or aggregating embeddings of words.
A major drawback to this approach is that bag-of-words models disregard word order and relationships between words that may serve as important information for classification.
Consider the following tweets that mention ``governor'':

\noindent $\ast$ [\textsc{Authority}] \textit{Love our \underline{governor}'s honesty \#njsandy}

\noindent $\ast$ [\textsc{Fairness}] \textit{Only 14 months till marriage \#Equality comes to NJ, when @CoryBooker is sworn in as next \underline{governor}}.

In the first tweet, two positive words ``love'' and ``honesty'' around ``governor'' obviously reflect the user's attitude towards him. In the second one, however, ``governor'' is not closely intertwined with other words and only modified by a neutral word ``next''. Because bag-of-words features ignore such context, the classifier may mistakenly assign Authority/Subversion to tweet 2 if ``governor'' is selected as a feature.

To address this issue, we experimented with various supervised learning models and found that the Recurrent Neural Network-based classifier with long short-term memory (LSTM)~\cite{hochreiter1997long} performed the best. LSTM is a specific Recurrent NN variant designed to better model long-term dependencies.
LSTM cells take as input a sequence of embeddings of words $\lbrace w_1, w_2, ..., w_l\rbrace$ in a tweet and output hidden states $\lbrace h_1, h_2, ..., h_l\rbrace$ in succession. We use the last output $h_l$, which is expected to encode information of the entire tweet, and handle it with a fully connected layer. Extra features including background knowledge are represented as separate vectors and processed using fully connected layers as well. Finally, we concatenate processed vectors and add a softmax layer on top for classification. To prevent overfitting, we apply dropout to outputs of the embedding, LSTM, and fully connected layers, and L2 regularization to the weight of the softmax layer. We train a separate classifier for each foundation and merge results from all classifiers.

\subsection{Textual Features}

In this paper, we use the following textual features.

\textbf{Word Embedding}:
Word embedding is a dense distributed representation which embeds words to a low-dimensional space to encode their semantic and syntactic information. We use $300$-dimensional Word2Vec embeddings trained on Google News\footnote{https://code.google.com/archive/p/word2vec}.
We also carried out experiments using unigram and/or bigram as features. As they didn't outperform word embedding and the latter is a common input to neural networks, we chose embeddings as the basic feature.

\textbf{Moral Foundation Dictionary}:
Linguistic Inquiry and Word Count (LIWC)~\cite{pennebaker2001linguistic} is a program that counts the proportion of words in different psychologically meaningful categories. Researchers have reported success applying LIWC to a range of social psychology problems~\cite{pennebaker2011secret,schwartz2013personality}.
In this work, we use Moral Foundation Dictionary~\cite{graham2009liberals}, a LIWC dictionary that contains $324$ foundation-supporting and foundation-violating words and word stems under $11$ categories. It can be regarded as a kind of moral-oriented prior knowledge, while it is not as rich as the knowledge we propose to utilize.

\section{Background Knowledge}

\begin{figure*}[!ht]
\centering
\includegraphics[width=\textwidth]{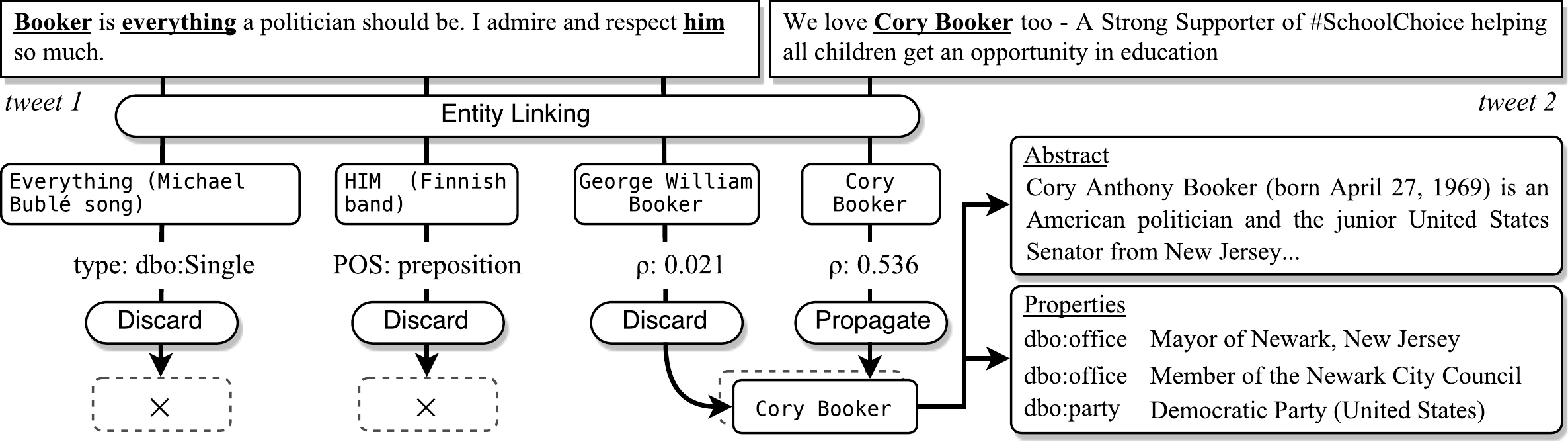}
\caption{Acquiring background knowledge.}
\label{figure:acquiring_background_knowledge}
\end{figure*}

Prior knowledge plays a critical role in how a human reader comprehends texts. As discussed above, background knowledge is also important in understanding expressions of moral concerns. For example, to perceive the Fairness/Cheating and Purity/Degradation-related moral concerns in the sentence ``\textit{we would also like to ban KKK}'', we need to know that ``KKK'' refers to \texttt{Ku Klux Klan}, hate groups opposing the Civil Rights Movement and same-sex marriage.

\subsection{Background Knowledge Acquisition}

To incorporate background knowledge, we apply entity linking to associate mentions with their referent entities.
Next, we develop a set of criteria to automatically remove or correct erroneous linking results based on their types, linking confidence scores, or part-of-speech tags. From the KB, we extract structured and unstructured information of remaining entities.
We will elaborate each step with the example illustrated in Figure~\ref{figure:acquiring_background_knowledge}.

\textbf{Entity linking}. First, we identify and link mentions to  entities in the KB using \textsc{TagMe}~\cite{ferragina2010fast}, a system developed to link mentions to pertinent Wikipedia pages. We choose this tool because most entity linkers are designed for formal texts such as news articles, while \textsc{TagMe} is intended for short texts and includes a special mode to handle hashtags, usernames, and URLs in tweets.
\textsc{TagMe} provides an open API\footnote{https://sobigdata.d4science.org/web/tagme/}, which returns a response including identified mentions, offsets, confidence scores, and Wikipedia titles. For tweet 1 in Figure~\ref{figure:acquiring_background_knowledge}, the linker identifies ``Booker'', ``everything'', and ``him'' and associates them with ``\texttt{George William Booker}'', ``\texttt{Everything (Michael Bubl\'e song)}'', and ``\texttt{HIM (Finnish band)}''.

\textbf{Result refinement}.
\textsc{TagMe} not only annotates capitalized phrases, thereby covering more mentions in poorly composed tweets at the cost of aggressively identifying some non-name words as mentions, such as ``everything'' and ``him'' in this example.
Additionally, the lack of information that contextualizes the mention stands an obstacle to entity disambiguation. For example, with the only clue - ``politician'' - in tweet 1, it is difficult to determine whether ``Booker'' refers to \texttt{George William Booker} or \texttt{Cory Booker} as both of them are politicians. To reduce these two types of errors, namely spurious annotations (``everything'' and ``him'') and linking errors (``Booker''$\rightarrow$\texttt{George William Booker}), we refine the results  based on the following attributes:

1. Linking confidence score. For each annotation, \textsc{TagMe} returns a confidence score ($\rho$) that estimates the linking quality. We remove entities with a low score ($<0.1$)
such as \texttt{George William Booker} ($\rho = 0.021$) in the example.

2. Type of entity. Under most circumstances, concepts incorrectly linked to non-name words are literary or musical work entities, such as songs and books, which are more possibly titled using common words. 
We collect all entity types in DBpedia\footnote{http://wiki.dbpedia.org/} and manually discard $113$ types.

3. Part-of-speech. In general, a single verb, adjective, adverb, pronoun, determiner, or preposition is unlikely to be a name. Thus, ``him'' acting as a pronoun in tweet 1 should not be marked as a name. We utilize a tweet-oriented part-of-speech tagger \cite{owoputi2013improved} to annotate the part-of-speech of each word. If no word in a mention matches any nominal tag, we will remove the associated entity from the results.

\textbf{Cross-document propagation}. In the previous step, we present rules to reduce spurious annotations and linking errors. For the latter case, however, our goal is to leverage prior knowledge rather than merely eliminating incorrect entities. Therefore, if the linker returns an annotation with a low score, we reject it and try to retrieve the referent entity from annotations of the same mention in other documents.
We make the following assumption: within a topic, when people mention the same name, they usually refer to the same entity. For example, in tweets regarding architecture, it is very likely that all mentions of ``Zaha'' refer to \texttt{Zaha Hadid}, an architect, instead of \texttt{Wilfried Zaha}, a footballer. Analogously, as it is difficult to determine the referent entity of ``Booker'' in tweet 1, we check annotations of other ``Booker''s in the entire corpus, find the most confident one (``Cory Booker'' in tweet 2$\rightarrow$\texttt{Cory Booker}, $\rho=0.536$), and use it as the entity of ``Booker'' in tweet 1.

\textbf{Knowledge extraction}.
Unlike human beings, machines still lack the ability to process and comprehend complicated information (e.g., a man carries an American flag upside down in an image in the \texttt{Westboro Baptist Church} Wikipedia page, which can be viewed as a political statement or an act of desecration and disrespect) or disregard information contributing little to moral value prediction (e.g., population of \texttt{New York} State). For this reason, we only derive two types of constructive knowledge that can be processed and utilized by existing techniques and are applicable to most entities from the KB as follows.

1. Entity abstract: a summary of an entity, which usually contains useful facts such as definition, office, party, and purpose.

2. Entity property: structured metadata and facts of an entity.
We obtain entity properties from DBpedia and
keep the following: \texttt{purpose}, \texttt{office}, \texttt{background}, \texttt{meaning}, \texttt{orderInOffice}, \texttt{seniority}, \texttt{title}, and \texttt{role}.

\subsection{Background Knowledge Incorporation}

Many facts in the retrieved background knowledge, however, are irrelevant to the prediction of moral values (e.g., term of office and education information of \texttt{Cory Booker}).  In addition, unlike tweets that are limited to $140$ characters, an abstract can either be very concise or contain up to hundreds of words. In order to extract discriminative information from the background knowledge, we design a pointwise mutual information (PMI)-based approach as follows.

We first merge the abstract and property values of an entity into a document. For example, the merged document of \texttt{Cory Booker} is ``\textit{Cory Anthony Booker (Born April 27, 1969) is an American politician ... Mayor of Newark, New Jersey. Democratic Party (United States)}''. After that, we calculate the document-based PMI with corpus level significance (cPMId)~\cite{damani2013improving} between each word in the document and the target foundation.
We rank all words with respect to their cPMId's and choose the top $k$ ($k = 100$ in our experiments).
Thus, we extract words strongly related to each moral foundation as features.

Lastly, we encode the background knowledge as a vector consisting of $k$ elements, each of which represents the term frequency of a selected word.
However, consider this case where ``hurt'' is a feature, while ``injury'' is the word used in a document. We will ignore ``injury'' although it is a synonym for ``hurt''. 
To encode the background knowledge in a ``softer'' and more generalizable way, we calculate the cosine similarity $\mathrm{sim}(u, w)$ between embeddings of feature $u$ and each word $w$. If $\mathrm{sim}(u, w)$ exceeds a chosen threshold ($0.6$ in this work), we regard $w$ as an occurrence of $u$.

\section{Experiments}

\subsection{Data Set}

For this work, we use a corpus of $4,191$ tweets randomly sampled from a larger corpus of 7 million Tweets containing hashtags relevant to Hurricane Sandy, a hurricane that caused major damage to the Eastern seaboard of the United States in 2012. All tweets included in these analyses were processed to strip user mentions, URLs, and punctuation.

To establish ground truth for our analyses, three trained annotators coded the $4,191$ sampled tweets\footnote{The data set and annotation guideline designed based on the Moral Foundation Theory will be published in a separate paper.}. Coder training consisted of multiple rounds of annotation and discussion. After completing training, annotators coded for the presence or absence of each moral foundation dimension. Additionally, tweets that contained no moral rhetoric were coded as ``Non-moral''.
Gold-standard classes for each tweet were then generated by taking the majority vote for each class across all three coders.
Each tweet can be annotated with more than one moral concern at the same time.

\begin{table}[!hbt]
\small
\centering
\setlength\tabcolsep{2pt}
\begin{tabularx}{0.48\textwidth}{|>{\hsize=1.7\hsize}X|>{\centering\arraybackslash\hsize=.6\hsize}X|>{\centering\arraybackslash\hsize=.7\hsize}X|>{\centering\arraybackslash\hsize=1\hsize}X|}
\hline
\textbf{Foundation} & \textbf{Positive} & \textbf{Negative} & \textbf{Pos:Neg} \\
\hline
Care/Harm             & 1,802 & 2,389 & 1:1.33 (0.75) \\
Fairness/Cheating     & 667   & 3,524 & 1:5.28 (0.19) \\
Loyalty/Betrayal      & 574   & 3,617 & 1:6.30 (0.16) \\
Authority/Subversion & 935   & 3,246 & 1:3.47 (0.29) \\
Purity/Degradation    & 159   & 4,032 & 1:25.4 (0.04) \\
Non-moral & 713   & 3,478 & 1:4.88 (0.21) \\
\hline
\end{tabularx}
\caption{Data set statistics. Note that ``positive'' and ``negative'' do not refer to virtue and vice of a foundation. Rather, they indicate whether moral concern on a foundation (e.g., Fairness/Cheating) is reflected in a tweet or not.}
\label{table:data}
\end{table}

Class frequency analyses of the coded corpus revealed considerable negative bias, such that the absence of each class occurred with greater frequency than its presence (See Table~\ref{table:data}). However, this is unsurprising, as there is no reason to expect half or even close to half of the texts in this corpus to evoke a given moral domain. Nonetheless, extreme imbalance like this can inhibit classifier performance by inducing classification bias and failing to sufficiently represent the population of the infrequent class. To account for this in our experiments, we up-sample positive classes to prevent bias toward the majority class.

To evaluate the annotation quality of this corpus, we measure inter-annotator agreement (IAA) using prevalence-adjusted bias-adjusted kappa (PABAK)~\cite{sim2015kappa}, which is suitable for imbalanced data. Based on the widely referenced standards for Kappa proposed in \cite{landis1977measurement}, IAA scores of this data set range from \textit{moderate} ($0.41-0.60$) to \textit{almost perfect} ($0.81-1.00$).

\subsection{Overall Results}

We evaluate our model with three feature sets: word embedding alone (E), the combination of word embedding and background knowledge (E+BK), and the combination of all features (E+BK+MFD). 
Model performance is evaluated using F-scores generated from 10-fold cross-validation in Figure~\ref{table:exp_knowledge}.

\begin{table}[!thb]
\small
\centering
\setlength\tabcolsep{2pt}
\newcolumntype{C}{>{\centering\arraybackslash\hsize=0.9\hsize}X}
\newcolumntype{S}{>{\centering\arraybackslash\hsize=0.8\hsize}X}
\begin{tabularx}{.48\textwidth}{|>{\hsize1.8\hsize}X|>{\centering\arraybackslash\hsize=0.6\hsize}X|>{\centering\arraybackslash\hsize=0.6\hsize}X|>{\centering\arraybackslash\hsize=1\hsize}X|}
\hline
\textbf{Foundation} & E & E+BK & E+BK+MFD \\
\hline
Care/Harm            & 81.2 & \textbf{82.3} & 81.9 \\
Fairness/Cheating    & 66.1 & 70.7 & \textbf{70.8} \\
Loyalty/Betrayal     & 47.2 & \textbf{50.3} & \textbf{50.3} \\
Authority/Subversion & 68.3 & 69.3 & \textbf{69.9}\\
Purity/Degradation  & 34.7 & \textbf{ 37.4} & 37.0 \\
Non-moral & 61.7 & \textbf{64.2} & 63.5 \\
\hline
\end{tabularx}
\caption{Overall results (\%, F-score). E, BK, and MFD represent embedding, background knowledge, and Moral Foundation Dictionary, respectively.}
\label{table:exp_knowledge}
\end{table}

Our experiment results provide evidence that integrating background knowledge into the representation of tweets improves detection of moral values. The following example demonstrates this process for a tweet which contains Authority/Subversion rhetoric: 

\noindent$\ast$ [\textsc{Authority}] \textit{Holy shit \underline{Chris Christie} is asking for federal funds Sounds like a self hating \underline{republican} to me hurricanesandy}

After linking ``Chris Christie'' and ``republican'' to \texttt{Chris Christie} and \texttt{Republican Party (United States)}, we know the former is the 55th Governor of New Jersey and the latter a major political party in the United States.
As our automatic approach selects politics-related words including ``governor'' and ``party'' as features,
such background knowledge effectively confirms the moral sentiment on Authority/Subversion in this tweet.

In another example:

\noindent$\ast$ [\textsc{Purity}] \textit{Hurricane Sandy is an opportunity for believers to embody the perfect peace \underline{Isaiah} 26 3 talks about as we trust in \underline{HIM} hurricanesandy}

Although we successfully link ``Isaiah'' and use the prior knowledge to correct the prediction, the linker fails to associate ``HIM'' with God, which illustrates the limitations of existing techniques. Humans are able to make a quick inference about the referent of ``HIM'' from its distinct uppercase form because pronouns referring to God are often capitalized or uppercased. In contrast, it is difficult for machines to distinguish different ``HIM''s (e.g., a common yet uppercased pronoun, a pronoun referring to God, the Finnish rock band, etc.), especially in poorly composed texts such as tweets.

It should also be noted that there seems to be a relationship between the prevalence of the positive class for a given dimension and performance for that dimension. For example, we observe that all models perform well on Care/Harm, for which the data is relatively balanced (See Table~\ref{table:data} and~\ref{table:exp_knowledge}), while they produce particularly low scores on the most imbalanced foundation, Purity/Degradation.

In addition, we observe that adding the Moral Foundation Dictionary does not further improve the performance if we have background knowledge. Our framework automates the extraction of the latter, hence saving much manual effort.

\subsection{Comparison with the Human Annotator}

While we have demonstrated the viability of our approach for classifying moral rhetoric, to truly evaluate the performance of these models it is necessary to compare them to human coder performance. To do this, we had a minimally trained fourth coder annotate a sample of $300$ tweets and used both the coder's annotations and the predictions from our model to predict moral concerns on these tweets. This enables us to compare the performance of the model to the performance of an independent human annotator.

\begin{table}[!htb]
\centering
\small
\setlength\tabcolsep{3pt}

\begin{tabularx}{0.48\textwidth}{|>{\hsize=1.4\hsize}X|>{\centering\arraybackslash\hsize=.8\hsize}X|>{\centering\arraybackslash\hsize=.8\hsize}X|}
\hline
\textbf{Foundation} & \textbf{4th Coder} & \textbf{Our Model} \\
\hline
Care/Harm            & 76.0 & 76.3 \\
Fairness/Cheating    & 76.6 & 72.3 \\
Loyalty/Betrayal     & 62.2 & 69.5 \\
Authority/Subversion & 68.5 & 67.8 \\
Purity/Degradation   & 61.8 & 54.8 \\
Non-moral & 77.9 & 69.2 \\
\hline
\end{tabularx}
\caption{A comparison of performance between human and our method (\%, F-score).}
\label{table:exp_subset}
\end{table}

On most categories, our model performs comparably to the human annotator (see Table~\ref{table:exp_subset}). Though, notably, the model again obtains a low score on Purity/Degradation. We also observe a large gap in the prediction of Non-moral, which may indicate that humans have a stronger ability to recognize tweets without moral content.

We also observe that although our model achieves comparable performance to the human annotator, the latter is superior in understanding deeper information in text to make inference. For example, in the following tweet:

\noindent$\ast$ [\textsc{Loyalty}] \textit{There needs to be a proper balance between individual responsibility and collective obligation Superstorm Sandy has shown us that}

Although ``individual responsibility'' and ``collective obligation'' are not typical words for Loyalty/Betrayal, a human reader is able to understand that the author's concern on this foundation is reflected when discussing the balance between ``individual responsibility'' and ``collective obligation''. The model, however, is unable to capture their relationship to make the correct prediction.

\subsection{Remaining Challenges}

Despite the effectiveness of our proposed model, we encounter some unsolved problems over the study period. We summarize the main remaining challenges as follows.

Tweets are often too short to provide contextual cues sufficient for entity disambiguation. For example, for the tweet ``\textit{Willard is a Frickin Lying Hypocrite}'', it is hard for the entity linking system to determine which entity ``Willard'' refers to. Additionally, tweets are often poorly composed and need to be normalized. People extensively use elements such as hashtags, abbreviations, slangs, and emoticons in tweets, which affects the performance of the entity linker and classifiers.

Further, knowledge from KBs is relatively static and limited. Consider the following tweet, ``\textit{Sandy could be God s answer to Obama letting his countrymen die in \underline{Benghazi} and then lying about it}''. The entity linker can easily link ``Benghazi'' to the \texttt{Benghazi} city. However, the real concerned knowledge is the attack against United States government facilities in Benghazi in 2012 instead of other facts, such as the population of the city, in the KB. To address this issue, we need to exploit more comprehensive knowledge of other types or from other sources, such as news and tweets.

Additionally, in this work, entity types to remove and property types to keep in the background knowledge extraction step are manually selected due to the limitation of data size. Manual selection may introduce individual biases and weaken generalization ability of the model on other corpora and domains. With enough occurrences of entity and property types, a number of automatic feature selection methods are applicable, such as mutual information, chi square, and information gain.

\section{Related Work}

Recently NLP techniques have been successfully applied to computational social science. Combined with social network analysis, textual content analysis has shown promise in applications such as prediction of moral value~\cite{Sagi2014qf,dehghani2016purity}, power~\cite{gomez2003centrality,PR2013,KR2016}, expertise~\cite{Horne2016}, leadership role~\cite{tyshchuk2013evolution}, personality~\cite{golbeck2011predicting,schwartz2013personality}, gender~\cite{burger2011discriminating,rao2010classifying}, hate speech~\cite{waseem2016hateful,nobata2016abusive}, and social interaction~\cite{althoff2014ask,tan2016winning}. This work has extensively studied textual (e.g., $n$-gram and LIWC) and structural features (e.g., Twitter relationships) on a variety of online platforms.

Nevertheless, to the best of our knowledge, our work is the first attempt to incorporate background knowledge through entity linking to enhance implicit content analysis in the area of computational social science. It should be noted that although there is a study on incorporating background knowledge into movie reviews classification by \citep{boghrati2015incorporating}, their ``background knowledge'' refers to articles describing the target movies, which act like the Moral Foundation Dictionary whereas are completely different from the background knowledge we actively extract from the knowledge base. 

\section{Conclusions and Future Work}

Moral value prediction is a critical task for predictingpsychological variables and events.
Using it as a case study, we demonstrate the importance of acquiring background knowledge for extracting implicit information through our new framework. Our framework can also be adapted for other implicit sentiment prediction tasks that are convertible to a multi-label classification problem, such as detecting personality types through text analysis~\cite{goldberg1990alternative}.

In the future, we will exploit more up-to-date background knowledge from wider sources such as news articles. We also will detect specific moral value holders and target issues associated with each moral concern (e.g., women's rights is the issue of the moral concern on Fairness/Cheating in ``\textit{oppression of women must be tackled}''). Moreover, we are interested in uniting moral value prediction with a variety of applications such as implicit community membership and leadership roles detection in social networks and event prediction.

We believe computational social science research can establish a bridge between NLP techniques and social science theories. We apply computational methods to analyze social phenomena supported by social theories, while more complex and accurate models can help theory adjudication in social science as well.

\bibliography{ijcnlp2017}
\bibliographystyle{ijcnlp2017}

\end{document}